\documentclass[]{rsos}

\usepackage{booktabs}

\usepackage{array}
\newcolumntype{L}[1]{>{\raggedright\let\newline\\\arraybackslash\hspace{0pt}}m{#1}}
\newcolumntype{C}[1]{>{\centering\let\newline\\\arraybackslash\hspace{0pt}}m{#1}}
\newcolumntype{R}[1]{>{\raggedleft\let\newline\\\arraybackslash\hspace{0pt}}m{#1}}
\newcommand{\specialcell}[2][c]{%
  \begin{tabular}[#1]{@{}l@{}}#2\end{tabular}}

\usepackage{blindtext}
\usepackage{helvet}
\usepackage{courier}
\usepackage{graphicx}
\usepackage{subfig}

\usepackage{balance}
\usepackage{soul,color}
\soulregister\cite7
\soulregister\ref7
\soulregister\pageref7
\usepackage{hyperref}

\usepackage[font=small,labelfont=bf]{caption}
\newcommand{\changed}[1]{#1}
\begin{document}

\title{Emo, Love, and God: Making Sense of {\it Urban Dictionary}, a Crowd-Sourced Online Dictionary}

\author{
Dong Nguyen$^{1,2}$, Barbara McGillivray$^{1,3}$, and Taha Yasseri$^{1,4}$}

\address{$^1$The Alan Turing Institute, London, UK.
\\$^2$Institute for Language, Cognition and Computation, School of Informatics, University of Edinburgh, Edinburgh, UK.
\\$^3$Theoretical and Applied Linguistics, Faculty of Modern and Medieval Languages, University of Cambridge, Cambridge, UK.\\$^4$Oxford Internet Institute, University of Oxford, Oxford, UK.
}

\subject{human-computer interaction}


\corres{Dong Nguyen\\
\email{dnguyen@turing.ac.uk}}

\begin{abstract}
The Internet facilitates large-scale collaborative projects \changed{and} the emergence of Web~2.0 platforms, where producers and  consumers of content unify, has drastically changed the information market. On the one hand, the promise of the "wisdom of the crowd" has inspired successful projects such as Wikipedia, which has become the primary source of crowd-based information in many languages. On the other hand, the decentralized and often un-monitored environment of such projects may make them susceptible to \changed{low quality content}. In this work, we focus on Urban Dictionary, a crowd-sourced online dictionary. We combine computational methods with qualitative annotation and  shed light on the overall features of Urban Dictionary in terms of growth, coverage and types of content.
We measure a high presence of opinion-focused entries, as opposed to the meaning-focused entries that we expect from traditional dictionaries. Furthermore, Urban Dictionary covers many informal, unfamiliar words as well as proper nouns.
\changed{Urban Dictionary also contains  offensive content, but highly offensive content tends to receive lower scores through the dictionary's voting system.
The low threshold to include new material in Urban Dictionary enables quick recording of new words and new meanings, but the resulting heterogeneous content can pose challenges in using Urban Dictionary as a source to study language innovation.}
\end{abstract}

\maketitle

%
%
%
%


\section{Introduction}

Contemporary information communication technologies open up new ways of cooperation leading to the emergence of large-scale crowd-sourced collaborative projects \cite{estelles2012towards}.
Examples of such projects are open software development \cite{dabbish2012social}, {\it citizen science} campaigns \cite{sauermann2015crowd}, and most notably Wikipedia \cite{doan2011crowdsourcing}.
All these projects are based on contributions from volunteers, often anonymous and non-experts. Although the success of most of these examples is beyond expectation, there are challenges and shortcomings to be considered as well.
In the case of Wikipedia for instance, inaccuracies \cite{giles2005internet}, edit wars and destructive interactions between contributors \cite{kittur2007he,yasseri2012dynamics}, and biases in coverage and content \cite{halavais2008analysis,samoilenko2014distorted}, are only a few to name among many undesirable aspects of the project that have been studied in detail.

The affordances of  Internet-mediated crowd-sourced platforms has also led to the emergence of crowd-sourced  online dictionaries.
Language is constantly changing. Over time, new words enter the lexicon, others become obsolete, and existing words acquire new meanings  (i.e. senses)\cite{labovsocial}. Dictionaries record new words and new meanings, are regularly updated, \changed{and sometimes used as a source to study language change \cite{siemund_2014}}. However, a new word or a new meaning needs to have enough evidence backing it up before it can enter a traditional dictionary. For example, \emph{selfie} was the Oxford dictionaries word of the year in 2013 and its frequency in the English language increased by 17,000\% in that year. Its first recorded use dates back to 2002,\footnote{\url{http://blog.oxforddictionaries.com/press-releases/oxforddictionaries-word-of-the-year-2013/}} but was only added to OxfordDictionaries.com in August 2013. 
Even though some of the traditional online dictionaries, such as Oxford Dictionaries\footnote{\url{https://www.oxforddictionaries.com}} or Macmillan Dictionary,\footnote{\url{https://www.macmillandictionary.com}} have considered implementing  crowdsourcing in their workflow \cite{abel-meyer_2013} \changed{(see \cite[p. 3-6]{rundell_2016} for a typology of crowdsourcing activities in lexicography)}, for most, they rely on professional lexicographers to select, design, and compile their entries.

Unlike traditional online dictionaries \changed{\cite[p. 11]{rundell_2016}},  the content in crowd-sourced  online dictionaries  comes from non-professional contributors and popular examples are Urban Dictionary\footnote{\url{https://www.urbandictionary.com/}} and Wiktionary \cite{meyer2012wiktionary}.\footnote{\url{https://en.wiktionary.org/}}
Collaborative online dictionaries are constantly updated and have a lower threshold for including new material compared to traditional dictionaries \changed{\cite[p. 2]{rundell_2016}}. Moreover, it has also been suggested that such dictionaries might be driving linguistic change, not only reflecting it \cite{creese_2013, creese_2017}. 
\changed{
Crowd-sourced dictionaries could potentially complement online sources such as Twitter, blogs and websites (e.g.,  \cite{10.1371/journal.pone.0113114,grieve_nini_guo_2017,Kerremans2011-KERTNI-2}) to study language innovation.}
However, such dictionaries are subject to spam and vandalism, as well as ``unspecific, incorrect, outdated, oversimplified, or overcomplicated descriptions'' \cite{abel-meyer_2013}. 
Another concern affecting such collaborative dictionaries is the question of whether their content reflects real language innovation, as opposed to the concerns of a specific community of users, their opinions, and generally neologisms and new word meanings that will not last in the language. 

\begin{figure}
\centering
 \includegraphics[scale=0.45]{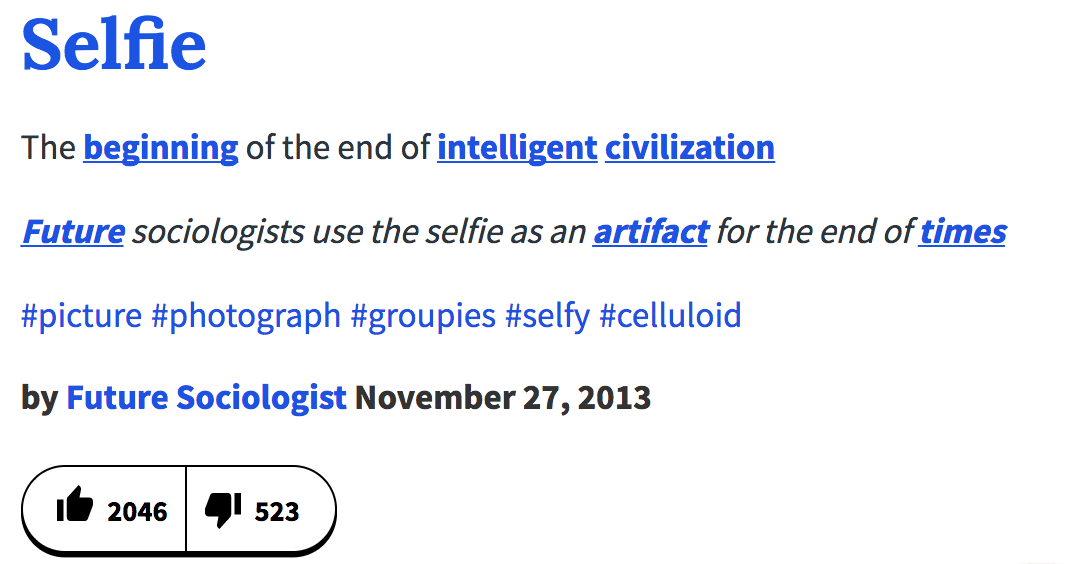}
 \caption{An Urban Dictionary entry for \emph{selfie}.}
 \label{selfie_example}
\end{figure}

\begin{figure}
\captionsetup{justification=centering}
    \centering
    \begin{minipage}{0.45\textwidth}
        \centering
    \includegraphics[width=0.85\textwidth]{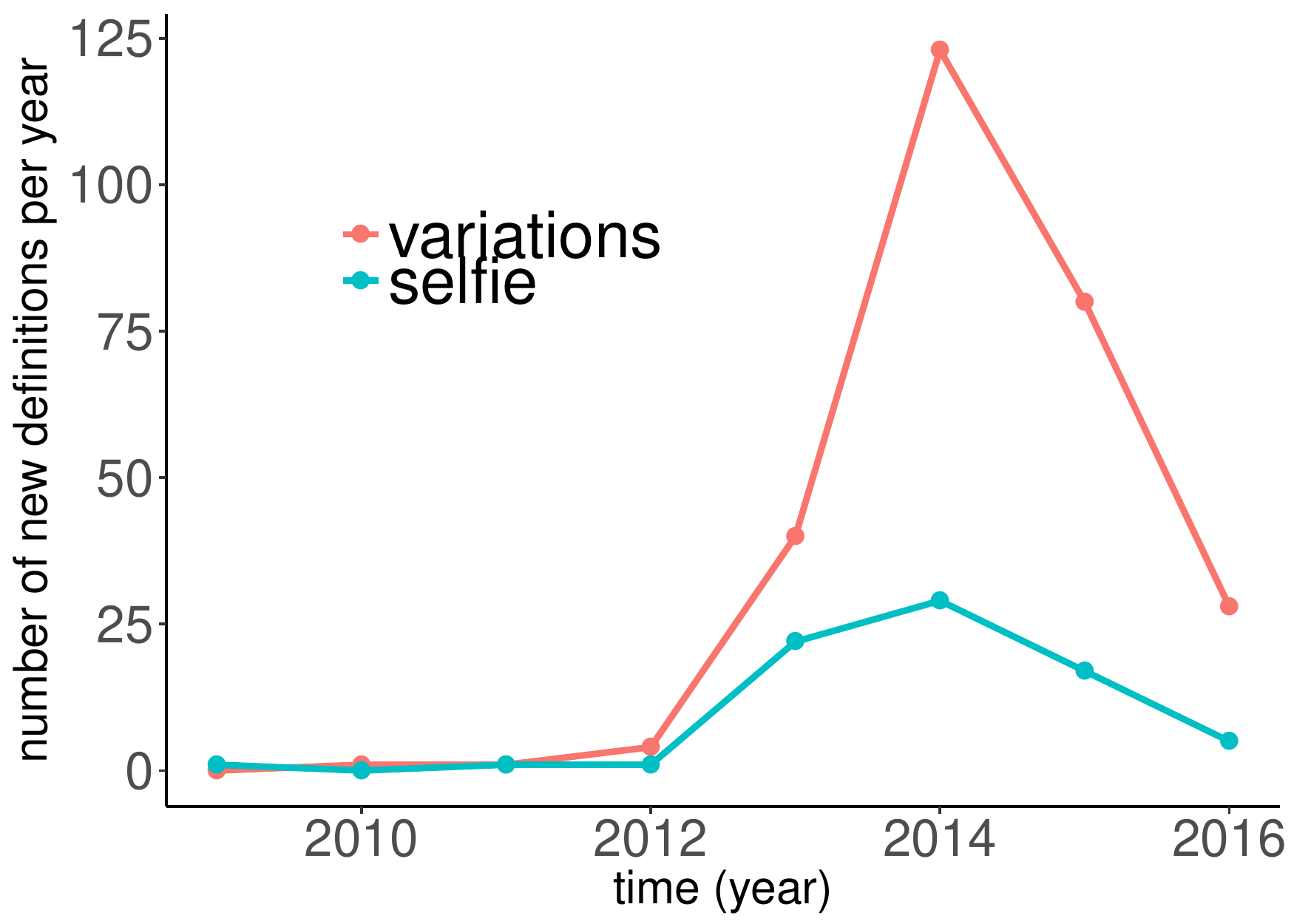}
      \caption{The number of new definitions for \emph{selfie} and its variations per year (Dec 1999 -- July 2016).}
      \label{selfie_time}
    \end{minipage}\hfill
    \begin{minipage}{0.45\textwidth}
        \centering
    \includegraphics[width=0.85\textwidth]{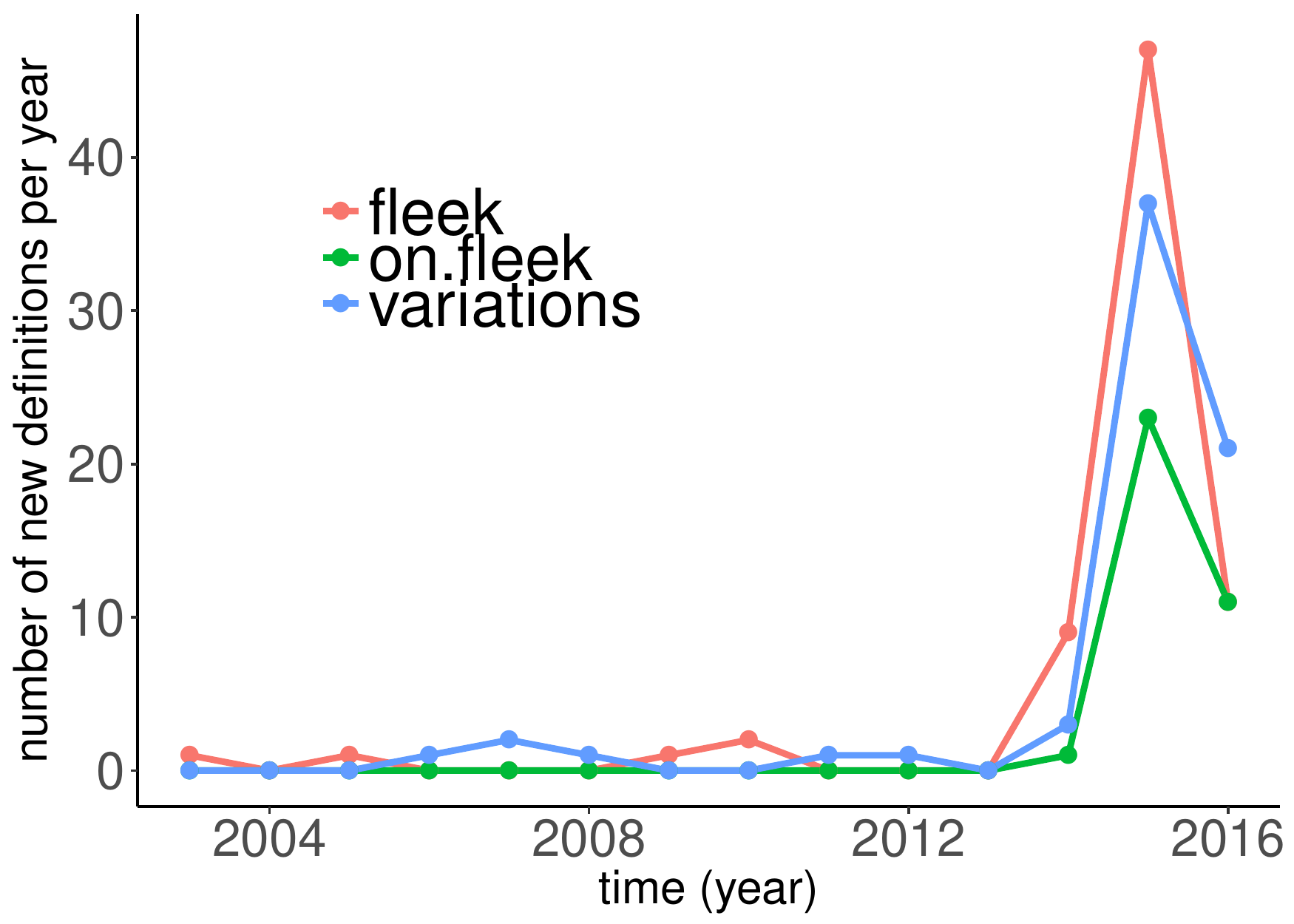}
     \caption{The number of new definitions for \emph{fleek} and \emph{on fleek} and other variations  per year (Dec 1999 -- July 2016).}
       \label{fleek_time}
     \end{minipage}
\end{figure}

This paper \changed{presents an explorative study of} Urban Dictionary (UD), an online crowd-sourced dictionary \changed{founded} in December 1999.  Users contribute by submitting an entry describing a word and a word might therefore have multiple entries. 
According to Aaron Peckham, its founder, ``\textit{People write really opinionated definitions and incorrect definitions. There are also ones that have poor spelling and poor grammar [$\ldots$] I think reading those makes definitions more entertaining and sometimes more accurate and honest than a heavily researched dictionary definition.}''\cite{tenore_2012}. 
 An  UD entry for \emph{selfie} is shown in Figure~\ref{selfie_example}, in which \emph{selfie} is defined as `\textit{The beginning of the end of intelligent civilization}.' and accompanied with an example usage `\textit{Future sociologists use the selfie as an artifact for the end of times}'. Furthermore, entries can contain tags (e.g., \emph{\#picture, \#photograph}). 
In total, Urban Dictionary contains 76 entries for \textit{selfie} (July 2016), the earliest submitted in 2009, and a range of variations (e.g., \textit{selfie-conscious, selfied, selfieing} and \textit{selfie-esteem}). Overall, there are 353 entries that describe a word (or phrase) containing the string \emph{selfie} (see Figure~\ref{selfie_time} for a plot over time). Figure~\ref{fleek_time} shows a similar plot for \emph{fleek} and \emph{on fleek}, a phrase that went viral in 2014. UD thus not only captures  new words rapidly, but it also captures the many variations that arise over time. 
Furthermore, the personal,  informal, and often offensive nature of the content in this popular site is \changed{different from the content typically found in}  both traditional dictionaries \changed{(see e.\,g. \cite[p. 3-4]{rundell_2016} and \cite[p. 7]{rundell_2016})} and more regulated collaborative dictionaries like Wiktionary. The status of UD as source of evidence for popular and current usage is widely recognized \cite{davis_2011, heaton_2010, Smith2011} and it has even been consulted in some legal cases \cite{ni_2017}. \changed{UD has also been used as a source to cross-check emerging word forms identified through Twitter \cite{grieve_nini_guo_2017}.}

\changed{Urban Dictionary has also been used} for the development of natural language processing systems that have to deal with informal language, non-standard language, and slang.  For example, UD has been consulted when building a text normalization system for Twitter \cite{beckley_2015} and it has been used to create more training data for a Twitter-specific sentiment lexicon \cite{tang_2014}.
In a recent study, Urban Dictionary is used to automatically generate explanations of  non-standard words \changed{and phrases
\cite{ni_2017}.}  

While Urban Dictionary seems to be a promising resource to record and analyze language innovation, so far little is known about the characteristics of its content. 
In this study we take the first step towards characterizing UD. 
\changed{
So far, UD has been featured in a few studies, but these qualitative analyses were based on a
small number of entries \cite{damaso_cotter_2007,Smith2011}.}
We study a complete snapshot \changed{(Dec 1999 -- July 2016)} of all the entries in the dictionary as well as selected samples using \changed{content analysis methods}. To the best of our knowledge, this is the first systematic study of Urban Dictionary at this \changed{scale.}

\section{Results}
\label{sec:results}
We start with presenting an overall picture of Urban Dictionary (Section 2 \ref{sec:overall_picture}), such as its growth and how content is distributed. \changed{Next, we compare its size to Wiktionary based on the number of headwords} (Section 2~\ref{sec:lexical_coverage}). 
We then present results based on two crowdsourcing experiments in which we analyze the types of content and the offensiveness of the entries (Section 2~\ref{sec:content_analysis}). Finally, we discuss how characteristics of the entries relate to their popularity on UD (Section 2~\ref{sec:content_popularity}).

\subsection{Overall picture}
\label{sec:overall_picture}
\begin{figure}
\captionsetup{justification=centering}
    \centering
    \begin{minipage}{\textwidth}
        \centering
    \includegraphics[width=0.35\textwidth]{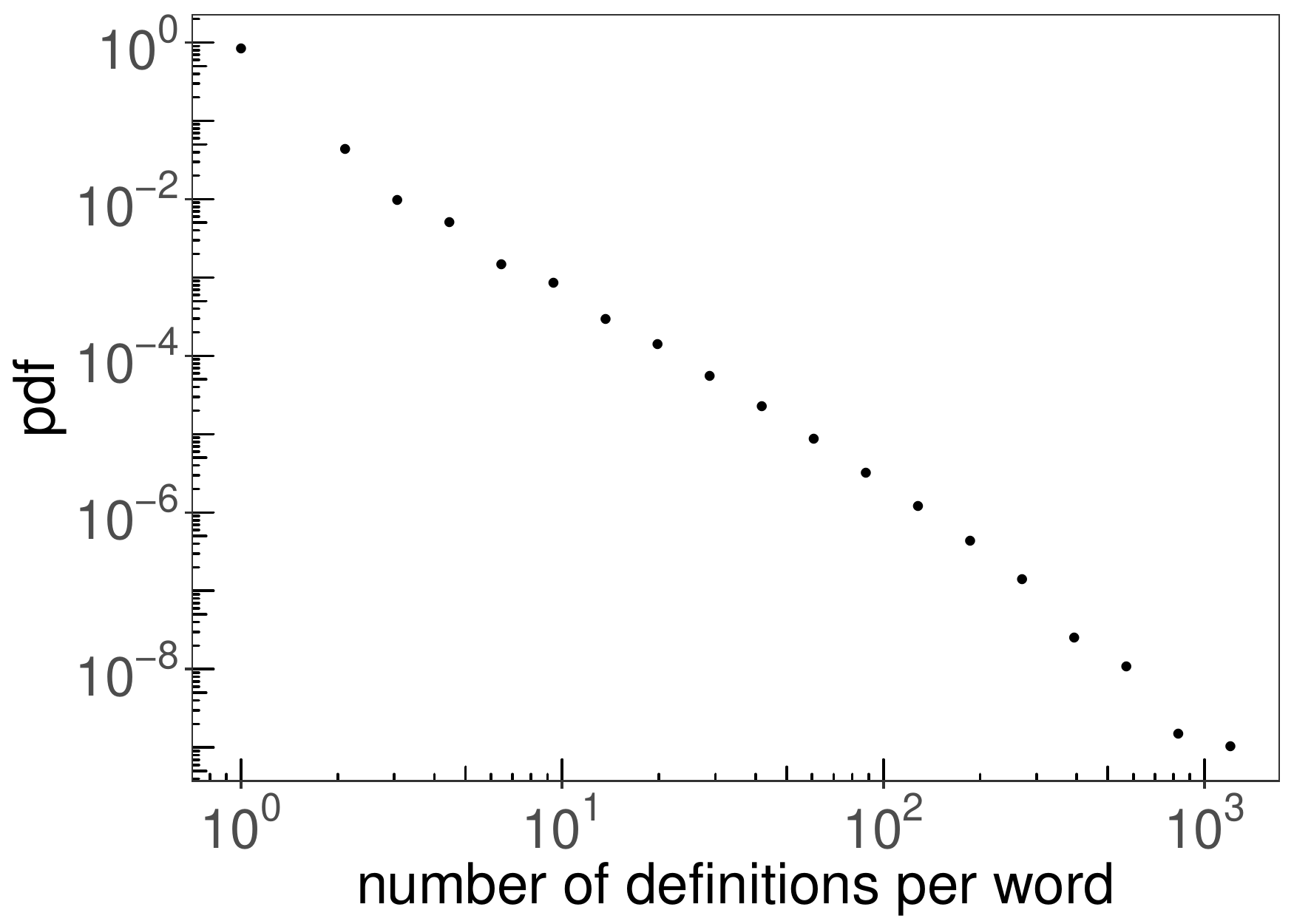}
    \includegraphics[width=0.35\textwidth]{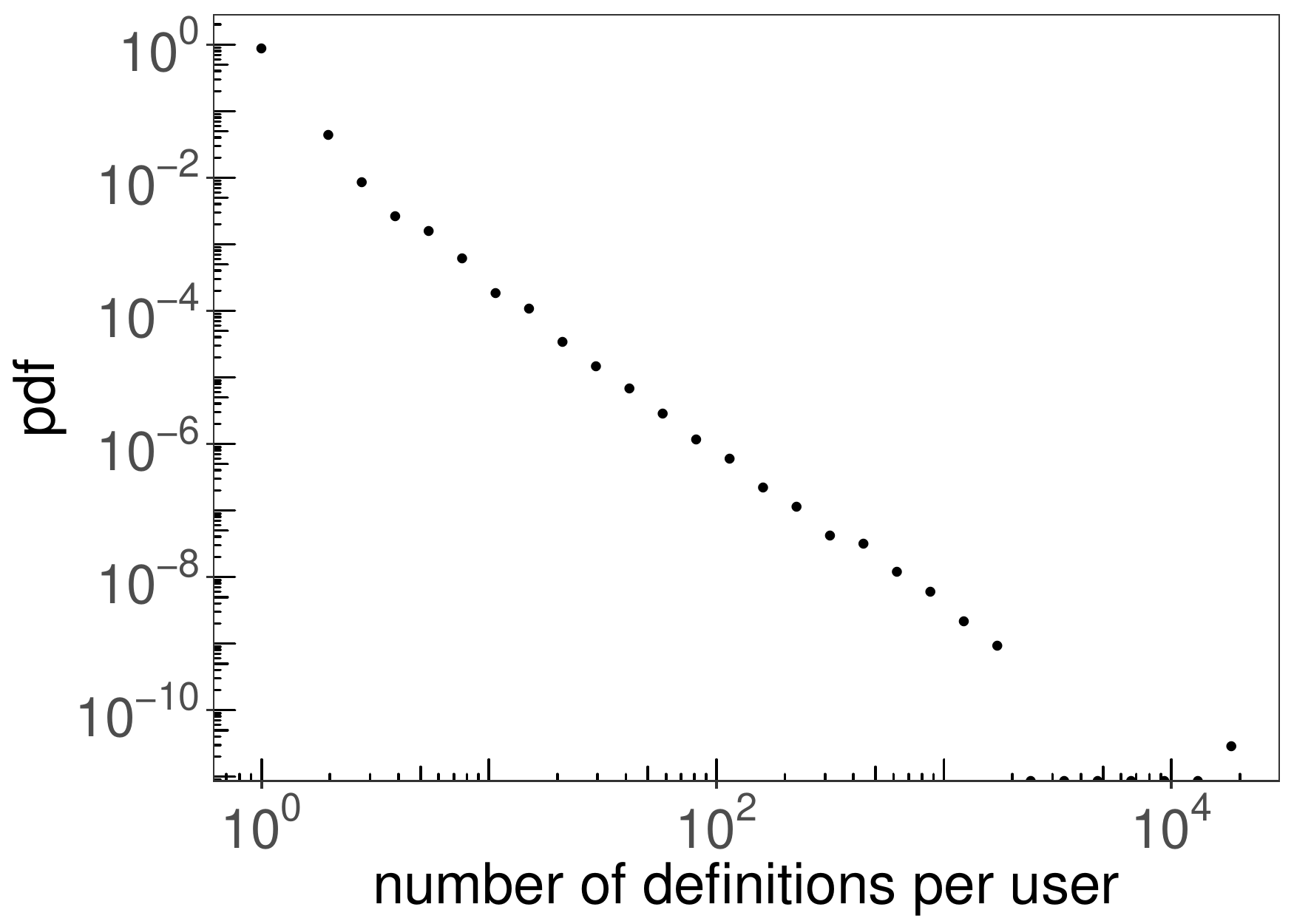}
      \caption{The probability density function of left: the number of definitions contributed to each headword and, right: the number of definitions contributed by each user of Urban Dictionary (logarithmic binning). Both axes are logarithmically scaled.}
       \label{def_pdf}
    \end{minipage}
\end{figure}

Since its inception in 1999, UD has had a rather steady growth. Figure~\ref{week} shows the number of new entries added each week. 
So far, UD has collected 1,620,438 headwords (after lower casing)\footnote{
We use `headword' to refer to the title under which a set of definitions appear. 
For example, in Wiktionary, the page about \emph{bank} covers different part of speech (e.g., noun and verb) as well as the different senses. 
In the context of UD, we use `entry' to refer to an individual content contribution (e.g., the combination of headword, definition, example text and tags submitted by a user).
Due to the heterogeneity in UD, we lower cased the headwords to calculate this statistic. This follows the interface of UD, which also does not match on case when grouping entries. } 
and 2,661,625 entries with an average of 1.643 entries per headword. 
However, as depicted in Figure~\ref{def_pdf}~(left), the distribution of the number of entries for each headword varies tremendously from one headword to another. While the majority of headwords have only one definition, there are headwords with more than 1,000 definitions. Table~\ref{top_terms_counts} reports the headwords with the largest number of definitions. \\

 \begin{minipage}{\textwidth}
  \begin{minipage}[b]{0.49\textwidth}
    \centering
     \includegraphics[width=0.9\textwidth]{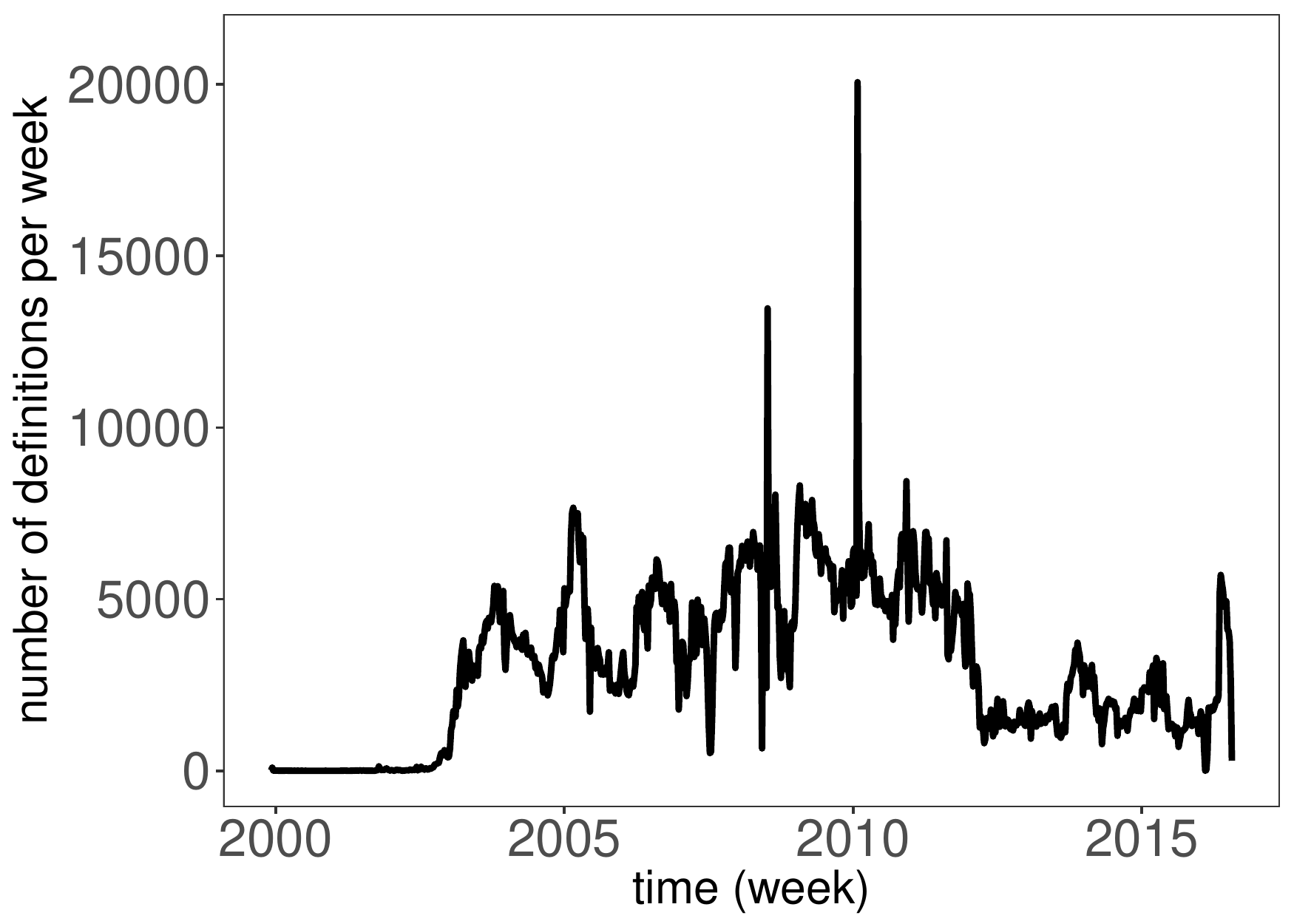}
      \captionof{figure}{Number of contributed definitions to Urban Dictionary per week since its inception in 1999.}
      \label{week}
  \end{minipage}
  \hfill
  \begin{minipage}[b]{0.49\textwidth}
    \centering
    \begin{tabular}{lr}
\toprule
\textbf{Headword} & \# \textbf{Definitions}\\
\midrule 
emo	&1,204\\
love	&1,140\\
god	&706\\
urban dictionary	&701\\
chode	&614\\
canada's history	&583\\
sex	&558\\
school	&555\\
cunt	&541\\
scene	&537\\
\bottomrule 
\end{tabular} 
\captionof{table}{Headwords with the most definitions.}
\label{top_terms_counts}
    \end{minipage}
  \end{minipage}

This fat-tailed, almost power-law distribution is not limited to the number of definitions per headword; the number of definitions contributed by each user follows a similar distribution, shown in Figure~\ref{def_pdf}~(right). 
The majority of users have contributed only once, while there are few power-users with more than 1,000 contributed definitions. 
These types of distributions are common in self-organized human systems, particularly similar crowd-based systems such as Wikipedia \cite{ortega2008inequality,yasseri2013value} or the citizen science projects Zooniverse \cite{sauermann2015crowd},  social media activity levels such as on Twitter \cite{huberman2008social}, or content sharing systems such as Reddit or Digg \changed{\cite{wu2007novelty}.} 

\begin{figure}
\captionsetup{justification=centering}
    \centering
    \begin{minipage}{\textwidth}
        \centering
    \includegraphics[width=0.35\textwidth]{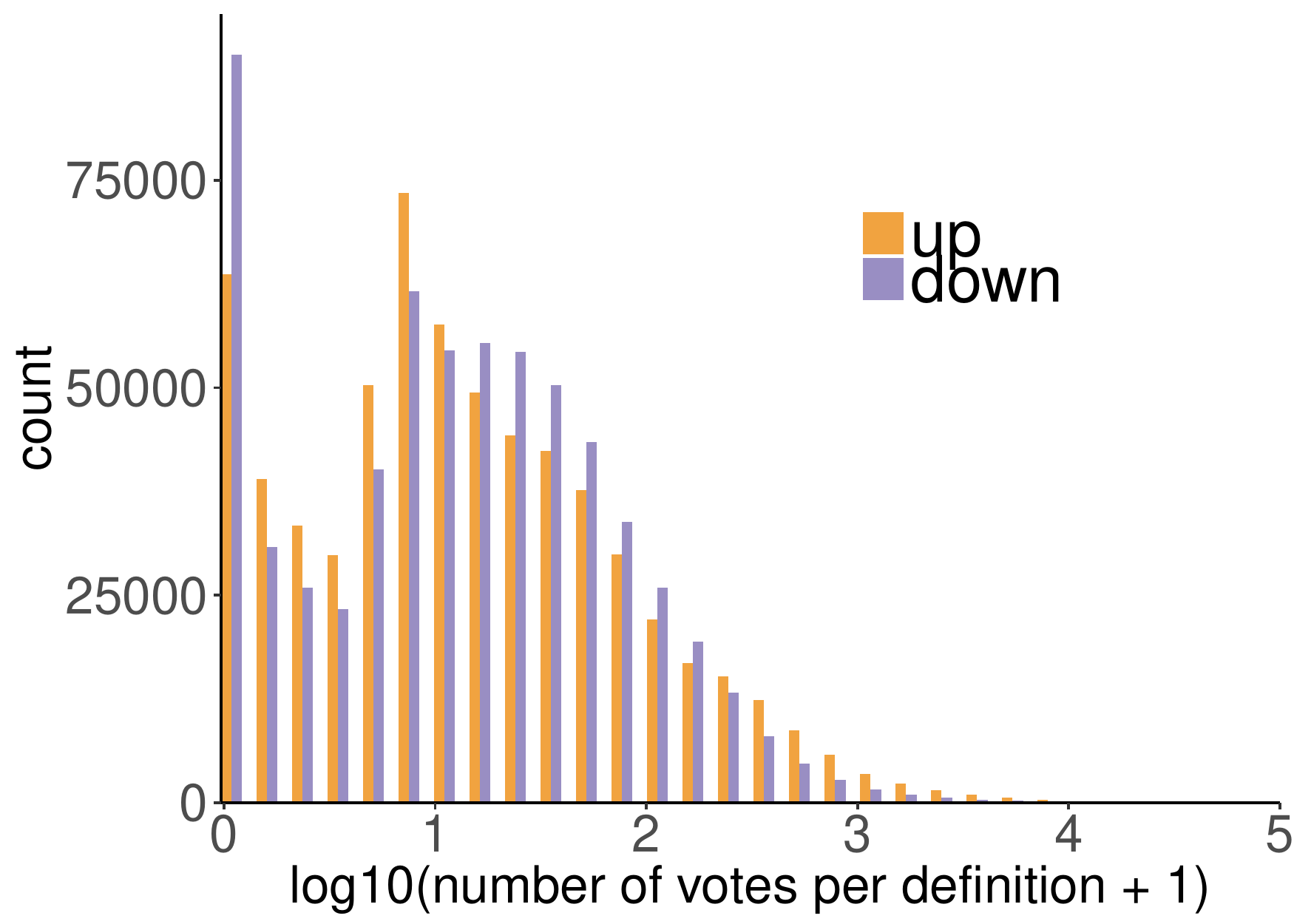}
    \includegraphics[width=0.30\textwidth]{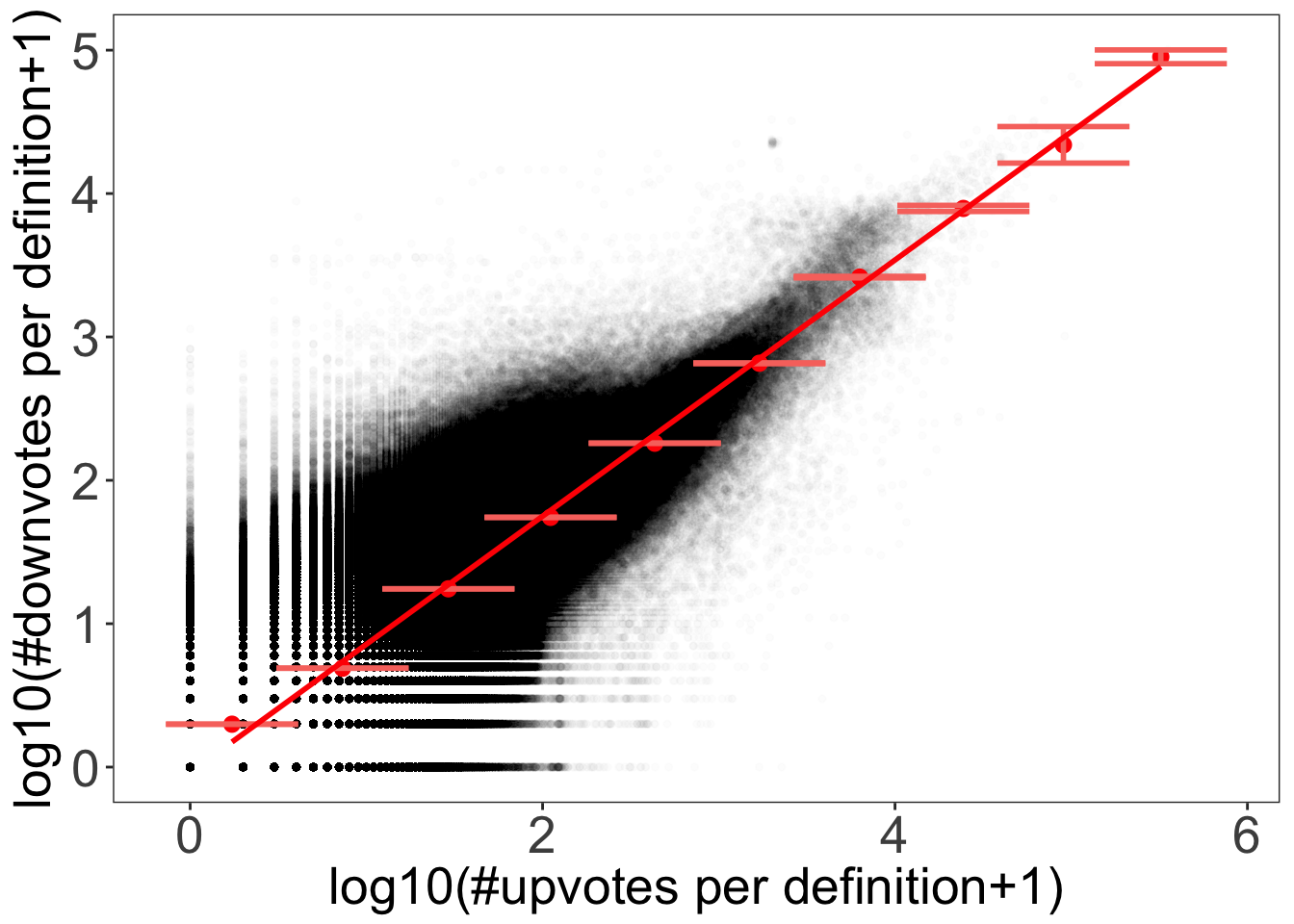}
    \includegraphics[width=0.30\textwidth]{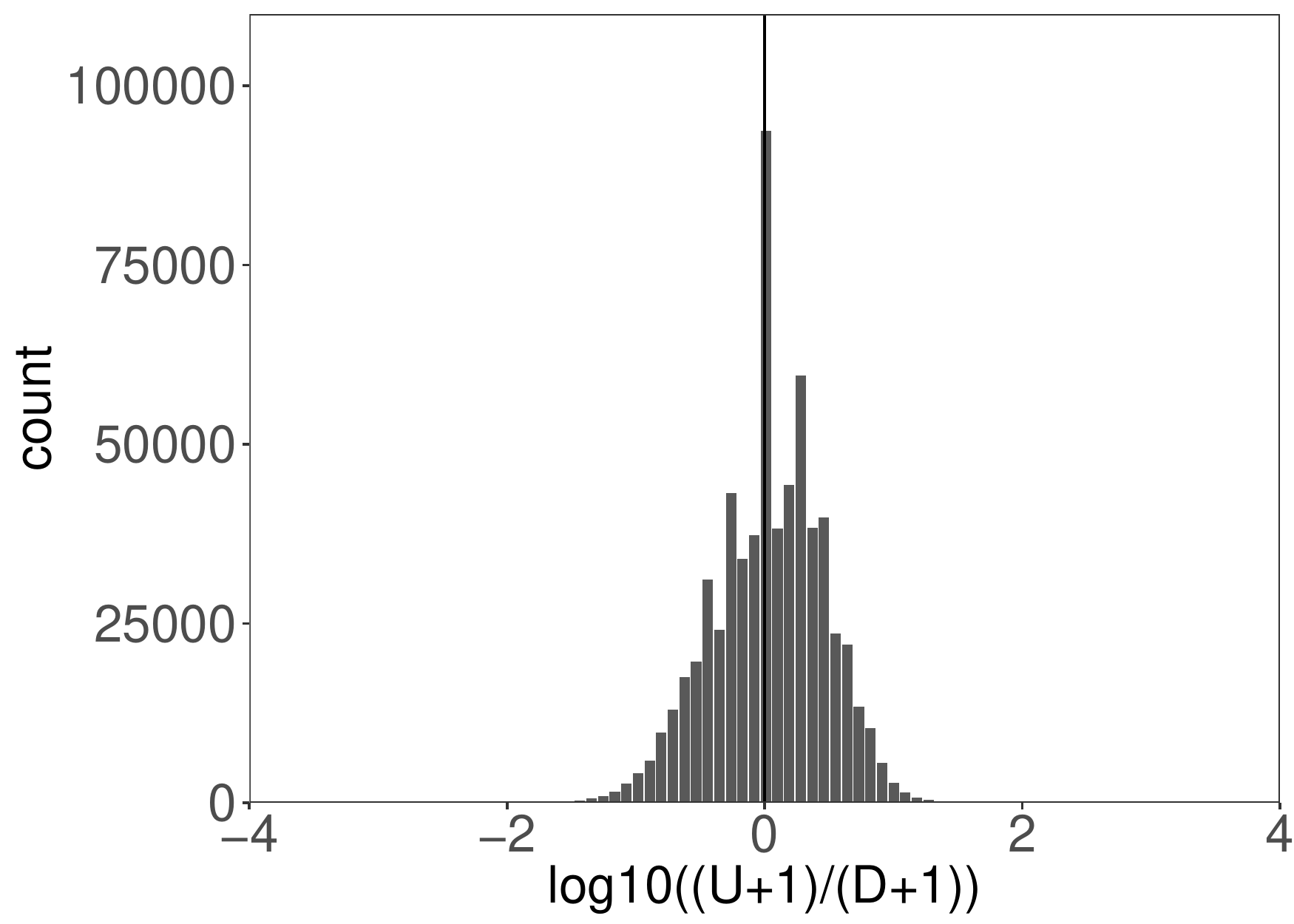}
      \caption{left: histogram of the number of votes of each definition, middle: scatter plot of the number of up votes and down votes that each definition has received, \changed{with error bars for bins and a fitted line}, right: the histogram of the ratio of up votes (U) to down votes (D) of each definition.}
       \label{vote}
    \end{minipage}
\end{figure}

A noteworthy feature of UD is that  users can express their evaluation of different definitions for each headword by up or down voting the definition. There is little to no guideline on "what a good definition is" in UD and users are supposed to judge the quality of the definitions based on their own subjective perception of how an urban dictionary should be. 
Figure~\ref{vote}~(left) shows the distribution of the number of up/down votes that each definition has received among all the definitions of all the headwords. A similar pattern is evident, in which many definitions have received very few votes (both up and down) and few definitions have many votes.  Figure~\ref{vote}~(middle) shows a scatter plot of the number of down votes versus the number of up votes for each definition. There is a striking correlation between the number of up and down votes for each definition which emphasizes  the role of visibility rather than quality in the number of votes. However, there seems to be a systematic deviation from a perfect correlation in which the number of up votes generally outperforms the number of down votes. This is more evident in Figure~\ref{vote}~(right), where the distribution of the ratio of up votes to down votes is shown. Evidently, there is a wide variation among the definitions with some having more than ten times more up votes than down votes and some the other way around. 

\subsection{\changed{Number of headwords}}
\label{sec:lexical_coverage}
We now compare \changed{the number of unique headwords in} Urban Dictionary to \changed{the number of unique headwords in Wiktionary,  another crowdsourced dictionary}.
\changed{Wiktionary  manifests a different policy than Urban Dictionary. The content in Wiktionary is created and maintained by 
administrators (selected by the community),  registered  users, and anonymous contributors \cite{meyer2012wiktionary}. In contrast to UD, there are many different mechanisms in  Wiktionary to ensure that the content adheres to the community guidelines. 
Each page is accompanied by a talk page, where users can discuss the content of the page and resolve any possible conflicts. Furthermore, in Wiktionary guidelines can be found for the structure and content of the entries}.
Capitalization is consistent and  content or headwords that do not meet the Wiktionary guidelines  are removed.  For example, while both UD and Wiktionary have misspelled headwords (e.g., \emph{beleive} for \emph{believe}),
\changed{Wiktionary guidelines state that only common misspellings should be  included while rare misspellings should be excluded\footnote{https://en.wiktionary.org/wiki/Wiktionary:Criteria\_for\_inclusion (17 February, 2018)}. In contrast, such guidelines are not present in UD.}
Wiktionary entries thus undergo a deeper level of curation.

Because of the inconsistent capitalization in UD, we experiment with three approaches to match the headwords between both dictionaries: no preprocessing, lower casing of all characters, and mixed.\footnote{The headword will be lower cased when the headword is all upper case or when the first character is upper case and the second character is lower case.} Table \ref{term_comparison_no_threshold} reports the result of this matching. The \changed{number of unique headwords in} UD is much higher and the lexical overlap is relatively low. 
Sometimes there is a match on the lexical level (i.e. the headwords match), but UD or Wiktionary cover different or additional meanings. For example, \emph{phased} is described in UD as 'something being done bit by bit -- in phases', a meaning also covered in Wiktionary. However, UD also describes several other meanings, including 'A word that is used when your asking if someone wants to fight.' and 'to be "buzzed." when you arent drunk, but arent sober.'.

\changed{Because there is little curation of UD content, there are many headwords that would not typically be included in a dictionary.}
 Examples include nick names and proper names (e.g. \emph{shaskank} defined as `Akshay Kaushik's nick name for his boyfriend Shashank.'; \emph{dan taylor}, defined as `A very wonderful man that cooks the best beef stew in the whole wide world. [\ldots]'), \changed{as well as 
informal spelling  (e.g., \emph{AYYYYYYYYYYYYYYYYYYY!!!!!!!!!!!!!!!!!!!!!!!!!!!!!!}) and 
  made-up words that actually no one uses (e.g. \emph{Emptybottleaphobia}\footnote{A Google search only returns 14 results, all of them containing the Urban Dictionary definition (17 Feb, 2018).}). Based on manual inspection, it seems that these are often headwords with only one entry}.

We therefore also perform a matching considering only headwords  from UD with at least two entries (Table \ref{term_comparison_at_least_two}). \changed{In this way, we use the number of entries as a crude proxy for whether the headword is of interest to a wider group of people.  Note that this filtering is not applied to Wiktionary, because each headword has only one page and headwords that do not match Wiktionary guidelines are already removed by the community. For example, an important criteri\changed{on} for inclusion in Wiktionary is that the term is \changed{reasonably widely} attested, e.g. has widespread use or is used in permanently recorded media\footnote{https://en.wiktionary.org/wiki/Wiktionary:Criteria\_for\_inclusion (17 February, 2018)}}. 
Compared to the first analysis, the difference is striking. \changed{In this comparison, the number of unique headwords in Wiktionary is higher than that of UD. }
From a manual inspection we see that many Wiktionary-specific headwords include  domain specific and encyclopaedic words (e.g., \emph{acacetins},  \emph{dramaturge} and \emph{shakespearean sonnets}), archaic words (e.g., \emph{unaffrighted}), as well as some commonly used words (e.g., \emph{deceptive}, \emph{e-voucher}). 
We also find that many of the popular UD headwords (i.e., headwords that have many entries) that are not covered in Wiktionary are proper nouns: The top five entries are \emph{canada's history},  \emph{justin bieber}, \emph{george w. bush}, \emph{runescape} and \emph{green day}.
In some cases, entries uniquely appearing in UD refer to words with genuine general coverage, such as \emph{loml} (in total 11 entries) defined as, e.g., `Acronym of "Love of My Life"' or \emph{broham} `a close buddy, compadre, smoking and/drinking buddy. a term of endearment between men to reaffirm heterosexuality.' (in total 18 entries). 

\begin{table}
\center
\begin{tabular}{lrrrrrr}
\toprule
& \multicolumn{2}{c}{\textbf{No processing}} & \multicolumn{2}{c}{\textbf{All lowercase}} & \multicolumn{2}{c}{\textbf{Mixed}}\\
\midrule
Overlap & 93,167 &(4\%) & 112,762& (5\%) &  108,361& (5\%)\\
Only UD & 1,698,812 &(72\%) & 1,507,675& (70\%) & 1,565,794 &(70\%)\\
Only Wiktionary & 569,787 &(24\%) & 540,641 &(25\%) & 546,263& (25\%)\\
\textbf{Total} & 2,361,766 && 2,161,078& & 2,220,418\\
\bottomrule
\end{tabular}
\caption{Headword comparison between UD and Wiktionary. The table reports the unique number of headwords in each category. No threshold was applied.}
\label{term_comparison_no_threshold}
\end{table}

\begin{table}
\center
\begin{tabular}{lrrrrrr}
\toprule
& \multicolumn{2}{c}{\textbf{No processing}} & \multicolumn{2}{c}{\textbf{All lowercase}} & \multicolumn{2}{c}{\textbf{Mixed}}\\
\midrule
Overlap &50,522 &(6\%) & 56,730 &(7\%)& 55,003 &(7\%)\\
Only UD &220,661 &(25\%)& 165,054 &(20\%)& 178,164 &(21\%)\\
Only Wiktionary & 612,432 &(69\%)& 596,673 &(73\%)& 599,621 &(72\%)\\
\textbf{Total} & 883,615 && 818,457 &&  832,788\\
\bottomrule
\end{tabular}
\caption{Headword comparison between UD and Wiktionary. The table reports the unique number of headwords in each category. Only UD headwords with at least two entries are included.}
\label{term_comparison_at_least_two}
\end{table}

\subsection{Content analysis}
\label{sec:content_analysis}
In this section we present our analyses on the different types of content as well as the offensiveness of the content in UD.

\subsubsection{Content type}
\changed{We now analyze several aspects of the content in UD that we expect to be different from content typically found in traditional dictionaries as well as Wiktionary. For example, manual inspection suggested that UD has a higher coverage of informal and infrequent words and of proper nouns (e.g., names of places or specific people). 
Many of the headwords are not covered in knowledge bases or encyclopedia\changed{s}. To characterize the data, we therefore annotated a sample of the data using crowdsourcing (see Data and methods). In order to limit the dominance of headwords with only one entry (which represent the majority of headwords in UD), the sample was created by taking headwords from each of the 11 frequency bins (see Table~\ref{bins} for details on the way the bins were created and sampled from). 
Note that the last two bins  are very small.
For each headword, we include up to three entries (top ranked, second ranked, and random based on up and down votes). 
Annotations were collected on the entry level and crowd workers were  shown the headword, definition and example.}

\begin{figure}
\captionsetup{justification=centering}
    \centering
      \begin{minipage}[t]{0.32\textwidth}
        \centering
 \includegraphics[width=0.92\textwidth]{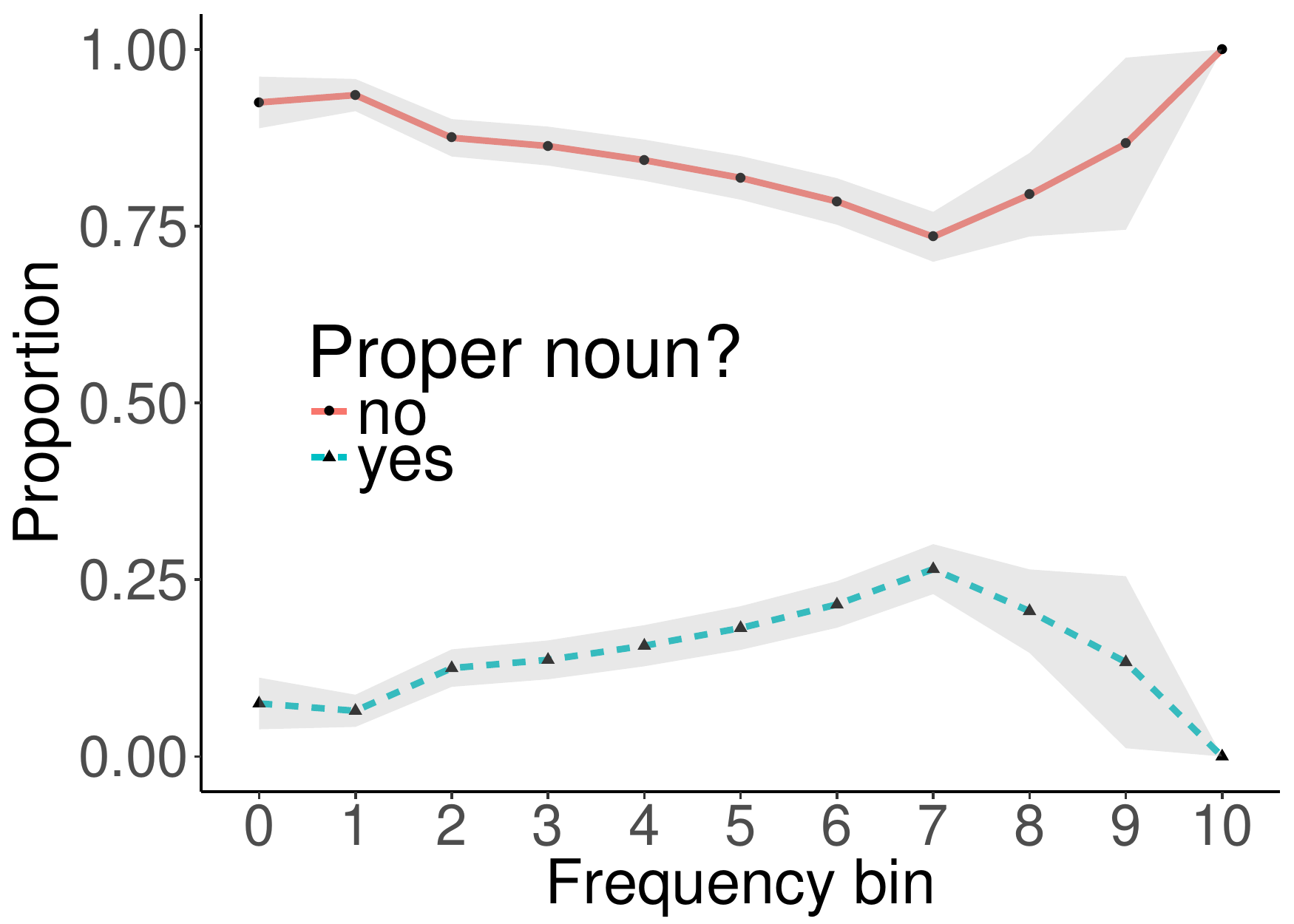}
      \caption{Proper nouns}
      \label{proper_nouns_proportion}
    \end{minipage}
    \begin{minipage}[t]{0.32\textwidth}
        \centering
    \includegraphics[width=0.92\textwidth]{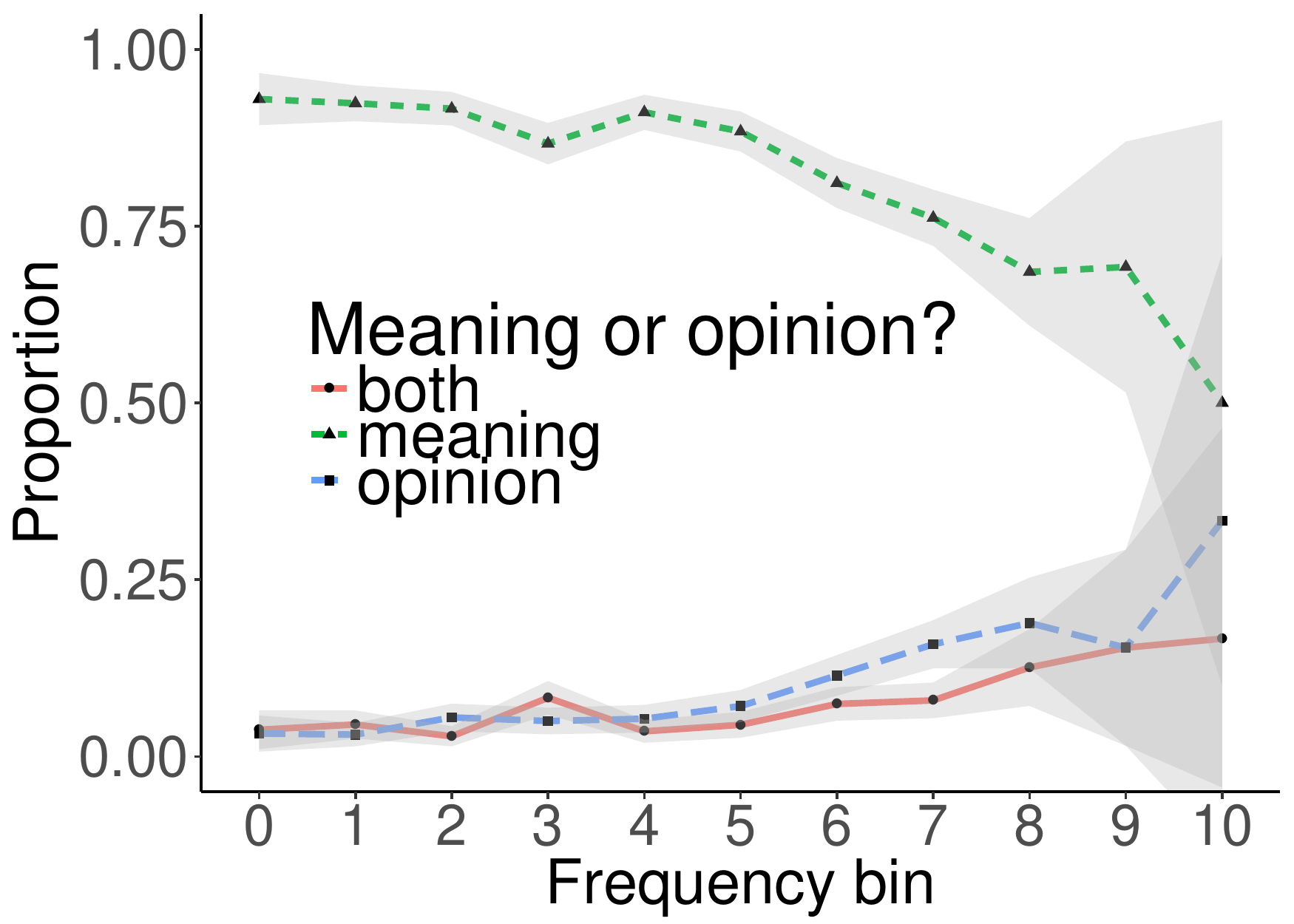}
      \caption{Meaning vs. opinions\\ (proper nouns were excluded)}
      \label{meaning_opinion_not_proper_nouns}
    \end{minipage}\hfill
    \begin{minipage}[t]{0.32\textwidth}
        \centering
    \includegraphics[width=0.92\textwidth]{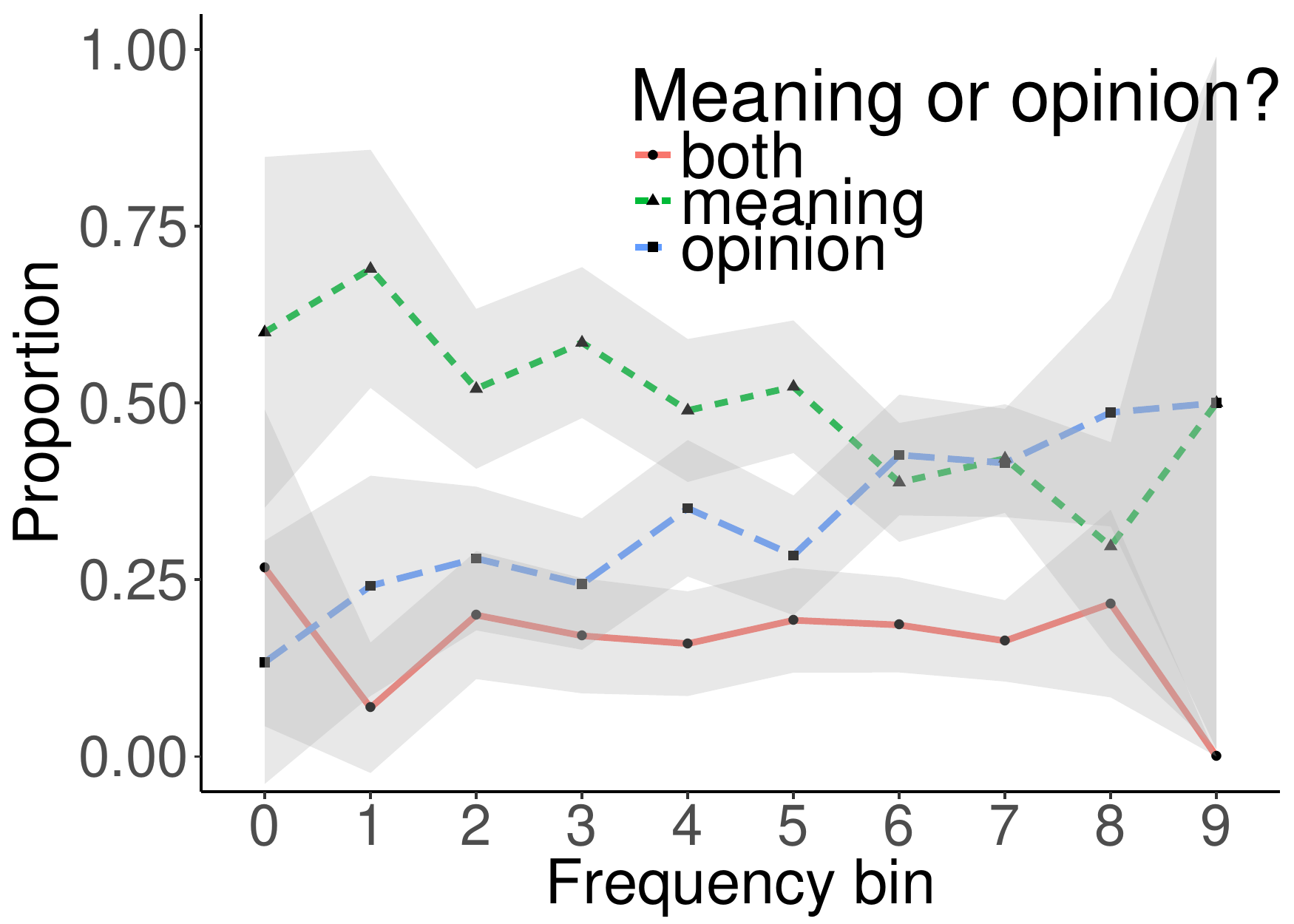}
      \caption{Meaning vs. opinions\\ (proper nouns entries only)}
       \label{meaning_opinion_proper_nouns}
     \end{minipage}\hfill
  
\end{figure}

\paragraph*{\changed{Proper nouns}}
\changed{Dictionaries are usually selective with including proper nouns  (e.g., names of places or individuals)} \changed{\cite[p. 77]{marconi_1990}.} 
In contrast, in UD many entries describe proper nouns. We therefore asked crowdworkers whether the entry described a proper noun (\textit{yes} or \textit{no}).
\changed{In our stratified  sample}, 16.4\% of the entries were annotated as being about a proper noun. Figure~\ref{proper_nouns_proportion} shows the fraction of proper nouns by frequency bin.

\paragraph*{\changed{Opinions}}
\changed{Most dictionaries strive towards objective content. For example, Wiktionary states `Avoid bias. Entries should be written from a neutral point of view, representing all usages fairly and sympathetically'\footnote{https://en.wiktionary.org/wiki/Wiktionary:Policies\_and\_guidelines  (16 Feb, 2018)}. } In contrast, the entries provided in UD do not always describe the meaning of a word, but they sometimes contain an opinion (e.g., \emph{beer} `Possibly the best thing ever to be invented ever. I MEAN IT.' or \emph{Bush} `A disgrace to America'). We therefore asked the crowdworkers whether the definition describes the meaning of the word,
 expresses a personal opinion, or
 both.
Figures \ref{meaning_opinion_not_proper_nouns} and \ref{meaning_opinion_proper_nouns} show the fraction of entries labeled as \emph{opinion}, \emph{meaning} or \emph{both}, separated according to whether they were annotated as describing proper nouns. In higher frequency bins, the fraction of entries marked as \emph{opinion} is higher. We also find that the number of entries marked as \emph{opinion} is  higher for  proper nouns. While most entries are marked as describing a \emph{meaning}, the considerable presence of opinions \changed{suggests} that the type of content in UD is different than in traditional dictionaries \changed{\cite[p. 3-4]{rundell_2016}}.

\begin{figure}
\captionsetup{justification=centering}
    \centering
    \begin{minipage}{0.45\textwidth}
        \centering
    \includegraphics[width=0.8\textwidth]{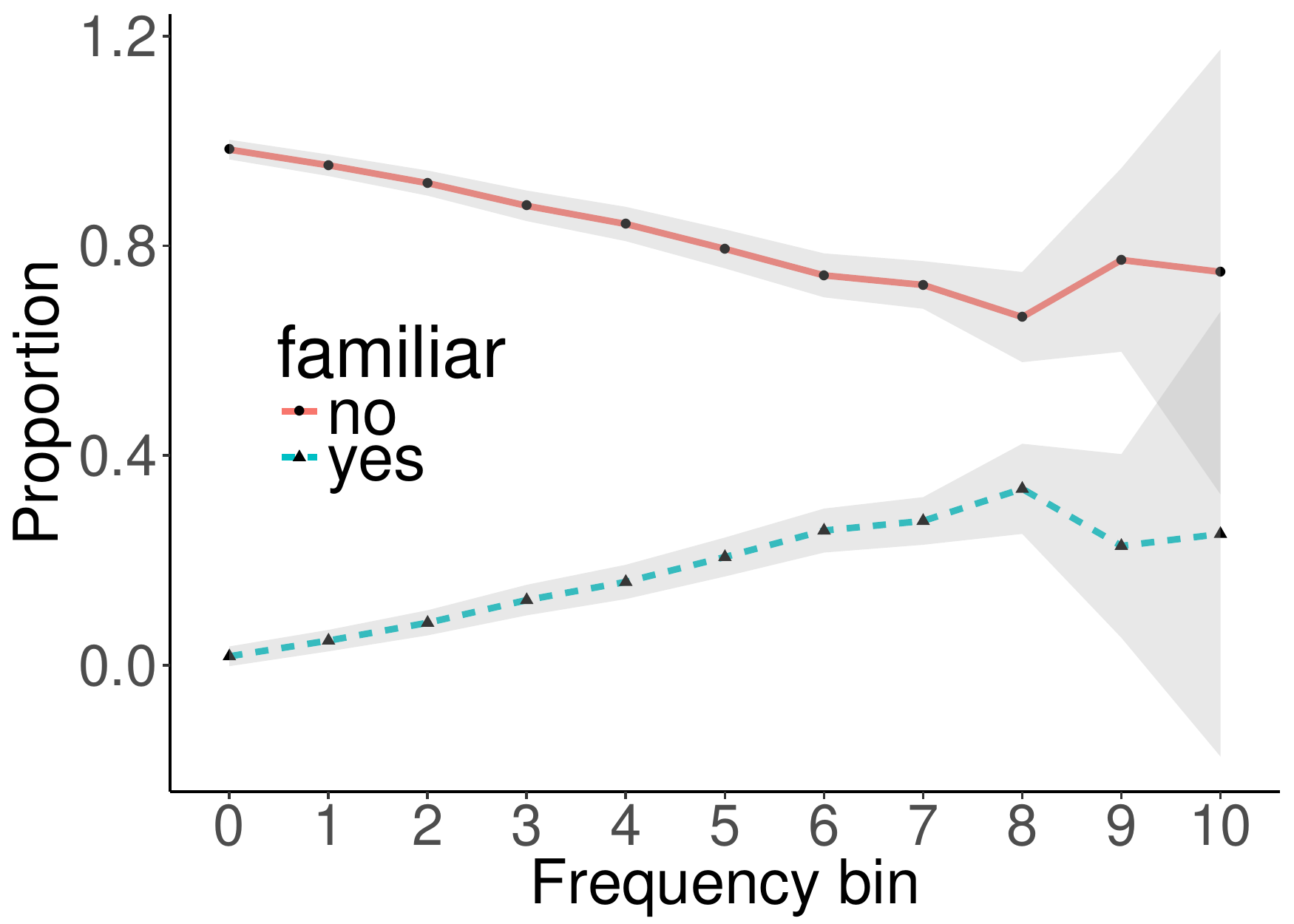}
      \caption{Familiarity\\ \changed{(proper nouns and opinion entries were excluded)}}
      \label{familiarity_plot}
    \end{minipage}\hfill
    \begin{minipage}{0.45\textwidth}
        \centering
    \includegraphics[width=0.8\textwidth]{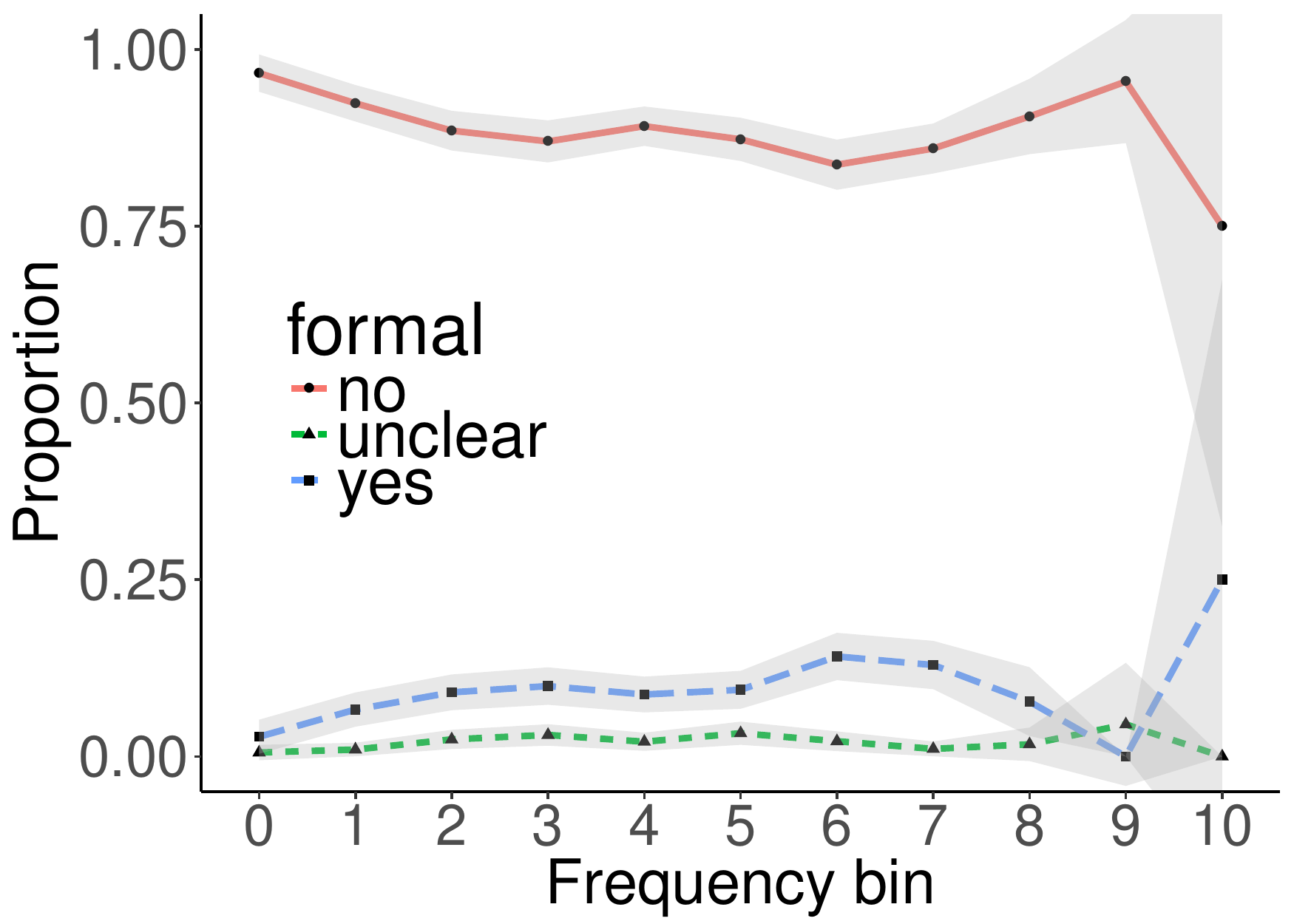}
      \caption{Formality\\ \changed{(proper nouns and opinion entries were excluded)}}
       \label{formality_plot}
    \end{minipage}\hfill

\end{figure}

\paragraph*{\changed{Familiarity}}
\changed{UD enables quick recording of new words and new meanings, many of them which may not have seen a widespread usage yet.  Furthermore, as discussed in the previous section, some entries are about made-up words or words that only concern a small community. 
In contrast, many dictionaries require that included headwords should be attested (i.e. have widespread use). 
These observations suggest that many definitions in UD may not be familiar to  people.
To quantify this, we asked crowdworkers whether they were familiar with the meaning of the word.}
The majority of the entries in UD were not familiar to the crowdworkers. Examples are common headwords with an uncommon meaning such as \emph{coffee} defined as `a person who is coughed upon' or \emph{shipwreck} 'The opposite of shipmate. 
A crew member who is an all round liability and as competent as a one legged man in an arse kicking competition.`,  as well as uncommon headwords and uncommon meanings (e.g., \emph{Once-A-Meeting} defined as `An annoying gathering of people for an hour or more once every pre-defined interval of time (ex: once a day).  
Once-A-Meetings could easily be circumvented by a simple phone call or e-mail but are instead used to validate a project managers position within the company.').
Figure \ref{familiarity_plot} shows that  in higher frequency bins, more definitions are marked as being \emph{familiar}, suggesting that the number of definitions per headword is indeed related to the general usage of a headword. 

\paragraph*{\changed{Formality}}
\changed{The focus of Urban dictionary on slang words \cite{peckham2009urban} means that many of the words are usually not appropriate in formal conversations, like a formal job interview. To quantify this, we asked crowdworkers whether the word in the described meaning can be used in a formal conversation.
As Figure \ref{formality_plot} shows, most of the words in their described meanings were indeed not appropriate for use in formal settings}. 

\subsubsection{Offensiveness}

Online platforms with user generated content are often susceptible to offensive content, which may be insulting, profane and/or harmful towards individuals as well as social groups \cite{Sood2012,waseem2017}. Furthermore,  
the existence of such content in platforms could signal to other users that such content is acceptable and impact  the social norms of the platform \cite{Sukumaran2011}. 
As a response, various online platforms have integrated different mechanisms to detect, report and remove inappropriate content. 
In contrast, regulation is minimal in UD and  one of \changed{its  characteristics}   is its often offensive content.

UD not only contains  offensive entries describing the meaning of offensive words, but there are also  offensive entries for non-offensive words (e.g., a definition describing \emph{women} as 'The root of all evil'). We note, however, that UD also contains  non-offensive definitions for  offensive words (e.g., \emph{asshole} defined as 'A person with no concept of boundaries, respect or common decency.'). To investigate how offensive content is distributed in UD, we ran a crowdsourcing task on CrowdFlower (see Data and methods for more details).  Workers were shown three definitions for the same headword, which they had to rank from the most to the least offensive.

\changed{We only included headwords with at least three definitions. In total, we obtained annotations for 1,322 headwords and thus 3,966 definitions. Out of these 1,322 headwords there are 326 headwords for which the majority of the workers agreed that \textit{none} of the definitions were offensive.}

Table~\ref{offensive_scores_meaning_opinion} reports the offensiveness  scores separated by whether the definitions describe a meaning, opinion or both.  An one-way ANOVA test indicates a slight significant difference (F(2, 3963) = 2.766, $p$<0.1).  A post hoc comparison using the Tukey test indeed indicates a slight significant difference between the scores of definitions describing a meaning and opinion ($p$<0.1). Thus, definitions stating an opinion tend to be ranked as more offensive compared to definitions describing a meaning.

Table~\ref{offensive_scores_formality} reports the offensiveness scores by formality. Definitions for words that were annotated as not being appropriate  for formal settings (based on their described meaning)   tend to  be ranked as being more offensive. An one-way ANOVA  confirms that the differences between the groups are highly significant
(F(2, 3963) = 22.72, $p$<0.001). 
Post hoc comparisons using the Tukey test
indicate significant differences between the formal and not formal categories ($p$<0.001), and between the unclear and not formal categories ($p$<0.05). 
We also find that definitions for which crowdworkers had indicated that they were familiar with the described meaning of the word tended to be perceived as less offensive (Table~\ref{offensive_scores_familiarity},  $p$ < 0.001 based on a t-test). 
We observe the same trends when we only consider definitions that describe a meaning. 
\\

 \begin{minipage}{\textwidth}
\begin{minipage}[b]{0.40\textwidth}
    \centering
   \begin{tabular}{lll}
\toprule
 \textbf{Type}&  \textbf{Avg. offensiveness}\\
 \midrule 
 both     &   2.025\\
 meaning    &    1.989\\
opinion   &     2.050\\
\bottomrule
\end{tabular} 
\captionof{table}{Average offensiveness rankings  (3=most offensive, 1=least offensive) by type of definition in UD entries.}
\label{offensive_scores_meaning_opinion}
\end{minipage}
\hspace{0.1\textwidth}
\begin{minipage}[b]{0.40\textwidth}
    \centering
  \begin{tabular}{lll}
\toprule
  \textbf{Formal?}&  \textbf{Avg. offensiveness}\\
 \midrule 
 no     &   2.031\\
 unclear    &    1.884\\
yes   &     1.873\\
\bottomrule
\end{tabular} 
\captionof{table}{Average offensiveness rankings  (3=most offensive, 1=least offensive) by formality in UD definitions.}
\label{offensive_scores_formality}
    \end{minipage}
    \end{minipage}

  \begin{table}
    \centering
    \begin{tabular}{lll}
\toprule
  \textbf{Familiar?}&  \textbf{Avg. offensiveness}\\
 \midrule 
   yes   &    1.915\\
 no  &     2.022 \\
\bottomrule
\end{tabular} 
\caption{Average offensiveness rankings  (3=most offensive, 1=least offensive) by familiarity in UD entries.}
\label{offensive_scores_familiarity}
 \end{table}

\subsection{Content and popularity}
\label{sec:content_popularity}
An important feature of UD is the voting mechanism that allows the users to express their evaluation of  entries by up or down voting \changed{them}. For a given headword, entries are ranked according to these votes and the top ranked one is labeled as \emph{top definition}. The votes thus drive the online visibility of entries, leading to the following implications. \changed{First, the top ranked entries are immediately visible when UD is consulted to look up the meaning of a headword. Many users might not browse the additional pages with   lower ranked entries.} Second, by users expressing their evaluation through votes, social norms are formed regarding what content is valued in UD. 

UD does not provide clear guidelines on "what a good definition is".  Various factors could  influence the up and down votes an entry receives, including whether the voter thinks the entry is offensive, informative, funny and whether the voter (dis)agrees with the expressed view. 
In this section we analyze how characteristics of the content as discussed in the previous section relate to the votes the entries receive. Because the number of up and down votes varies highly depending on the popularity of the headword, we perform the analysis based on the rankings of entries (top ranked, second ranked, and random) instead of the absolute number of up and down votes. Only headwords with at least three entries are included.

Table~\ref{characterization_votes_meaning_opinion} shows the distribution of opinion-based versus meaning-based definitions  separated by whether the headwords are annotated as proper nouns by the crowdworkers. The proportion of definitions that are annotated as opinions is much higher for proper nouns, which is consistent with our previous analysis. However, among the top ranked definitions for proper nouns, the proportion of opinions is lower (but n.s.).

Table~\ref{characterization_votes_familar_formal_offensiveness} characterizes the entries by formality and familiarity. We discard proper nouns and entries marked as opinion, since it is less clear what formality and familiarity mean in these contexts. We find that the top ranked definitions tend to be more familiar (${\chi}^2$ (2, N = 2991) = 15.385, p $<$0.001) and more appropriate for formal settings (but n.s.).

\begin{table}[h!]
\center
\begin{tabular}{llllll}
\toprule
& \multicolumn{3}{c}{\textbf{Opinion or meaning?}}\\ 
&\textbf{both}&\textbf{meaning}&\textbf{opinion}\\
\midrule
\multicolumn{4}{c}{\textit{No proper nouns (n=3,268)}}\\
top ranked & 0.055&0.852&0.094\\
second ranked & 0.074&0.850&0.076\\
random & 0.051& 0.864&0.084\\
\multicolumn{4}{c}{\textit{Proper nouns (n=698)}}\\
top ranked & 0.172&0.481&0.347\\
second ranked & 0.169&0.477&0.354\\
random & 0.190&0.444&0.366\\
\bottomrule
\end{tabular}
\caption{Characterization of UD entries based on votes. The table reports the proportions of opinion-based versus meaning-based definitions in each of the ranking groups.}
\label{characterization_votes_meaning_opinion}
\end{table}

\begin{table}[h!]
\center
\begin{tabular}{l|ll|lll|l}
\toprule
&\multicolumn{2}{c|}{\textbf{Familiar?}} &\multicolumn{3}{c|}{\textbf{Formal?}} & \textbf{Offensiveness} \\
&\textbf{no}&\textbf{yes}&\textbf{no}&\textbf{unclear}&\textbf{yes} & \textbf{avg. ranking}\\
\midrule
top ranked &  0.799&0.201     & 0.855&0.026& 0.119 & 1.950\\
second ranked & 0.807&0.193  &0.876&0.023& 0.101 & 1.966\\
random & 0.861&0.139            &0.894  &0.020&0.086 & 2.107\\

\bottomrule
\end{tabular}
\caption{Familiarity, formality and offensiveness of UD definitions across rankings based on votes. Definitions for proper nouns and definitions annotated as opinions are not included. The table reports the proportions in each of the rankings for familiarity and formality and the average ranking for offensiveness (3=most offensive, 1=least offensive); n=2,991.}
\label{characterization_votes_familar_formal_offensiveness}
\end{table}

Table~\ref{characterization_votes_familar_formal_offensiveness} also reports the average offensiveness ranking of the definitions separated by their popularity (again, discarding proper nouns and entries marked as opinions). 
The difference in rankings between  top ranked and second ranked definitions is minimal, but random definitions are  more often ranked as being more offensive.
A one-way ANOVA test  confirms that the differences between the groups are highly significant
(F(2, 2988) = 22.07, $p$<0.001). 
Post hoc comparisons using the Tukey test
indicate significant differences between the random and top ranked, and random and second ranked definitions ($p$<0.001). 
A similar trend is observed when we consider all definitions 
(F(2, 3963) = 34.87, $p$<0.001). 
Thus, although  \changed{UD contains  offensive content}, very offensive definitions do tend to be ranked lower through the voting system. However, the small difference in scores between the groups indicates that offensiveness only plays a small role in the up and down votes a definition receives.

To analyze the different factors jointly, we fit an 
ordinal regression model  (Table~\ref{orm_ranking}) using the \emph{ordinal} R library based on definitions that were annotated as not being an opinion and not describing proper nouns. 
We find that familiarity and offensiveness indeed have a significant effect. More familiar and less offensive definitions tend to have a higher ranking.
Similar trends in coefficients were observed with fitting logistic regression models when dichotomizing the ranking variable.

\begin{table}[!htbp] \centering 
\small
  \label{} 
\begin{tabular}{@{\extracolsep{5pt}}lD{.}{.}{-3} } 
\toprule
 \multicolumn{2}{r}{\textbf{Dependent variable: ranking}} \\ 
\midrule
Familiar (yes) &  -0.255^{***}$ $(0.096) \\ 
Formal (unclear) &  -0.133$\hspace{0.165in} $(0.226)\\
Formal (yes) &-0.073$\hspace{0.165in} $(0.123)\\
Offensiveness & 0.335^{***}$ $(0.059) \\ 
\midrule 
Observations & \multicolumn{1}{c}{2,991} \\ 
Log likelihood & -3262.19\\
AIC & 6536.38\\
\bottomrule\end{tabular} 
  \caption{Ordinal regression results. The dependent variable is the ranking:
  top ranked (0), second ranked (1) or a random rank (2).
  $^{***}$p$<$0.01.} 

\label{orm_ranking}
\end{table}

\section{Discussion and conclusion}
In this article, we have studied a complete snapshot (1999--2016) of Urban Dictionary to shed light on the characteristics of its content. 
\changed{We} found that most contributors of UD only added one entry and very few added a high number of entries. 
Moreover, we found a number of skewed distributions, which need to be taken into account whenever performing analyses on the UD data. 
Very few headwords have a high number of entries, while the majority have only one entry. Similarly, few entries are highly popular (i.e. they collected a high number of votes). We also found a strong correlation between the number of up and down votes for each entry, illustrating the importance of visibility on the votes an entry receives. 

The lexical content of UD is radically different from that of Wiktionary, another crowdsourced, but more highly moderated dictionary. In general, we can say that the overlap between the two dictionaries is  small. 
Considering all unique UD  headwords that are not found in Wiktionary, we found that this number is almost three times  the number of  headwords that uniquely occur in Wiktionary. However, if we exclude words with only one definition in UD (which tend to 
be infrequent or idiosyncratic words), we found the opposite pattern, with Wiktionary-only headwords amounting to almost three times the UD-only headwords. 

Our analyses based on crowdsourced annotations showed more details on the specific characteristics of UD content. 
In particular, we measured a high presence of opinion-focused entries, as opposed to the meaning-focused entries that we expect from traditional dictionaries. In addition, many entries in UD describe proper nouns. 
The crowdworkers were not familiar with most of the definitions presented to them and many words (and their described meaning) were found  not  to be appropriate for formal settings.

\changed{Urban Dictionary captures} many infrequent, informal words and it also contains \changed{ offensive} content, but highly offensive definitions  \changed{tend to get} ranked lower through the voting system.  The high content  heterogeneity in UD could mean that, depending on the goal,  \changed{considerable} effort is needed to filter and process the data (e.g., the removal of opinions) compared to when traditional dictionaries are used. 
We also found that words with more definitions tended to be more familiar to crowdworkers, suggesting that UD content does reflect broader trends in language use to some extent.

There are several directions of future work that we aim to explore. 
\changed{We have compared the lexical overlap with Wiktionary in terms of headwords.} As future work, we plan to extend the current study  by performing a deeper semantic analysis and by  comparing UD with other non-crowdsourced dictionaries. 
Furthermore, we plan to extend the current study by comparing the content in Urban Dictionary with language use in social media to advance our understanding of the extent to which UD reflects broader trends in language use.

\section{Data and methods}
\label{sec:data}

\subsection{Data collection}
\subsubsection{Urban Dictionary}
We crawled UD in July 2016.
First, the definitions were collected by crawling the `browse' pages of UD and by following the `next' links. After collecting the list of words,  the definitions themselves were crawled
directly after (between July 23 and July 29, 2016). We did not make use of the API, since the API restricted the maximum number of definitions returned to ten for each word.

\subsubsection{Wiktionary}
We downloaded the Wiktionary dump of the English language edition  of 20 July, 2016, so that the date matched our crawling process. To parse Wiktionary, we made use of code available through ConceptNet 5.2.2 \cite{speer2012representing}. Pages in the English Wiktionary edition can also include sections describing other languages (e.g., the page about \emph{boot} contains an entry describing the meaning of \emph{boot} in the Dutch language (`boat')). We only considered  the English sections in this study.

\subsection{Crowdsourcing}
Most headwords in UD have only one entry, and therefore these headwords would dominate a random sample. Because such headwords tend to be uncommon, a random sample would not be able to give us much insight into the overall content of UD.
We therefore sampled the headwords according to the number of their entries. For each headword (after lower casing), we counted the number of entries and place the headword in a frequency bin (after taking a log base 2 transformation).
For each bin, we randomly sampled up to 200 headwords. For each  sampled headword, 
we included the top two highest scoring entries (scored according to the number of thumbs up minus the number of thumbs down)  and another random entry. In total we sampled 4,465 entries (Table \ref{bins}).

\begin{table}
\begin{tabular}{rlllllllllll}
\toprule 
\textbf{Frequency bin (log 2)}&  0 &  1  & 2  & 3  & 4 &  5  & 6 & 7 &  8 &  9&  10 \\
\textbf{\#definitions} & 200 &449& 600& 600 &600& 600 &600& 600 &180 & 30 &  6 \\
\bottomrule 
\end{tabular} 
\caption{Statistics of the sampled definitions}
\label{bins}
\end{table}

We collected the annotations using CrowdFlower. The quality was ensured using test questions and by restricting the contributors to quality levels two and three and the countries Australia, Canada, Ireland, New Zealand, UK, and the USA. We marked the crowdsourcing tasks as containing explicit content, so that the tasks were only sent to contributors that accepted to work with such content.

\subsubsection{Content Type}
 For each task, we collected three judgements. The workers were paid \$0.03 per judgement. 
We collected 13,395 judgements from a total of 201 workers. The median number of judgements per worker is 76. Workers were shown the headword, definition and example.
The crowdworkers were asked the following questions (options for answers are displayed in italic font):
\begin{itemize}
\item  \textbf{Q1}: Is this word a proper noun, for example, a name used for an individual person (like Mark), place (like Paris), or organization (like Starbucks, Apple)? \textit{yes}, \textit{no}
\item  \textbf{Q2}: The definition:
 \textit{describes the meaning of the word,
 expresses a personal opinion,
 both}
 \item \textbf{Q3}: Were you familiar with this meaning of the word before reading this definition?
If you are familiar with this word but NOT with this meaning, then please select no. Example: If you are familiar with the meaning of the word `cat' as the animal, but the definition describes cat as `A person, usually male and generally considered or thought to be cool.' and you are not familiar with this meaning, select no: \textit{yes,
 no}
 \item \textbf{Q4}:
Can this word in the described meaning be used in a formal conversation?
Examples of formal settings are a formal job interview, meeting an important person, or court of law. Examples of informal settings are chatting with close friends or family:
\textit{ yes,
 no,
 unclear}
\end{itemize}

\paragraph*{Agreement}
For each definition we have three judgements. We calculate Fleiss' kappa (using the \emph{irr} package in R) and the pairwise agreement (Table~\ref{agreement_annotation}). The agreement for the first question, asking whether the word is a proper noun, is the highest. 
In general the agreement is low, due to the difficulty of the task. For example, in these cases all three workers answered differently to the question whether the definition described a meaning or an opinion: \emph{AR-15} defined as`{AR does NOT stand for Assault Rifle}' and \emph{Law School} defined as `Where you go for to school for four years after college to learn to become a lawyer. In these four years, you will work your butt off every day, slog through endless amounts of reading, suffer through so much writing, and after you graduate, you don't get to call yourself "doctor".'.
We merge the answers for each question by taking the majority vote. We use `\emph{both}' for Q2 and `\emph{unclear}' for Q4 if there was no majority.

\begin{table}
\center
\begin{tabular}{llll}
\toprule
& \textbf{Fleiss' kappa} & \textbf{Pairwise agreement}\\
\midrule
Q1: Proper noun (yes, no)& 0.379& 0.806\\
Q2: Meaning or opinion? (meaning, opinion, both)& 0.207  & 0.691\\
Q3: Familiar (yes, no) &0.206  & 0.713\\
Q4: Formal (yes, no, unclear)& 0.207 &0.712\\
\bottomrule
\end{tabular} 
\caption{Agreement statistics}
\label{agreement_annotation}
\end{table}

\subsubsection{Offensiveness}

We experimented with different \changed{pilot} setups in which we asked workers to annotate the level and type of offensiveness for individual definitions. However, we found that this led to confusion and disagreement among the crowdworkers. For example, an offensive word can be described in  a non-offensive way and a non-offensive word can be described in an offensive way.
\changed{Furthermore, people have different thresholds of what they consider to be offensive, making it challenging to ask for a binary judgement}.
In the final setup, we therefore showed the  sampled definitions for the \textit{same} word and asked workers to rank the definitions according to their offensiveness, with 1 being the most offensive and 3 being the least offensive. 
\changed{Even if workers have different thresholds of what they consider offensive, they could still agree when being asked 
to rank  the definitions}. Indeed, we found that this led to a higher agreement. Note that in this article, we have reversed the ratings (3=most offensive, 1=least offensive) for a more intuitive presentation of the results.
Workers were also asked to indicate whether they considered all definitions equally offensive, equally non-offensive, or none. 
For each task, we collected five judgements. We paid \$0.04 per judgement. 
We collected 6,610 judgements from a total of 158 workers (median number of judgements per worker: 44).
Table \ref{offensive_ranking_examples} provides examples for two words (\emph{goosed} and \emph{dad}) and their ratings.
\begin{table}
\begin{tabular}{lll}
\toprule
\textbf{Word} & \textbf{Definition} & \textbf{Ratings}\\
\midrule
 & \specialcell{\emph{Def. 1} Old school definition: to pinch someone's buttocks, hopefully the\\ opposite sex, but hey, you take what you get. Always associated in my\\ mind with a British accent....}	 & 2, 2, 2, 2, 2 \\ 
goosed & 	\specialcell{\emph{Def. 2} adj. 1. a feeling of overwhelmedness 2. a feeling of frusteration \\3. a feeling of joy 4. all emotions easily substituted by the word \\5. the new ``owned"} & 1, 1, 1, 1, 1\\
& \specialcell{\emph{Def. 3} To apply pressure on one's taint (or space between genitalia and \\anus), preferably of the opposite sex!	} & 3, 3, 3, 3, 3\\
\midrule
& \emph{Def. 1} the one who knocked-up your mom & 2, 2, 2, 2, 3\\
dad & \specialcell{\emph{Def. 2} The parent that takes the most shit. Sure, if you had a shitty father,\\ then go ahead and bitch, but not all of us did. Some of us had great fathers,\\ who really loved us, and weren't assholes. Honestly, if you could see\\ how much damage a mother could do to one's self esteem, you wouldn't \\even place so much blame on ``dear old dad"} & 3, 3, 3, 3, 2\\
& \specialcell{\emph{Def. 3} The replacement name for "bro" to call your best friend of whom\\ you have a fatherly bond} & 1, 1, 1, 1, 1\\
\bottomrule 
\end{tabular} 
\caption{Examples of annotated definitions for offensiveness (3=most offensive, 1=least offensive).}
\label{offensive_ranking_examples}
\end{table}

\paragraph*{Agreement} We calculate agreement using  Kendall's W (also called Kendall's coefficient of concordance), which ranges from 0 (no agreement) to 1 (complete agreement). We calculate Kendall's W for each word separately.
The average value of Kendall's W is 0.511 (standard deviation = 0.303). If we exclude words for which a worker indicated that the definitions were equal in terms of offensiveness, the value increases to 0.714 (standard deviation = 0.238).

\section*{Ethics statement}
In this study we employ crowdsourcing to collect annotations.
The tasks were marked as containing  explicit content, so that the tasks were only visible to contributors that accepted to work with such content. The tasks also explicitly mentioned that the results will be used for scientific research ('\textit{By participating you agree that these results will be used for scientific research}.').
We closely monitored the crowdsourcing tasks and contributor satisfaction was consistently high. 

\section*{Data accessibility}
The datasets supporting this article are available upon request.

\section*{Competing interests}
The authors declare no competing interests.

\section*{Authors' contributions}
DN collected and analyzed the data, participated in the design of the study, and drafted the manuscript; BM participated in the design of the study, analysed the data and drafted the manuscript; TY conceived the study, analyzed the data, and helped draft the manuscript. All authors gave final approval for publication.


\section*{Funding}
This work was supported by The Alan Turing Institute under the EPSRC grant EP/N510129/1.
DN was supported by Turing award TU/A/000006 and BM by Turing award TU/A/000010 (RG88751). The crowdsourcing data collection was supported with an Alan Turing Institute seed funding grant (SF024).

\sloppy
\emergencystretch 1.5em
\bibliographystyle{vancouver}
\bibliography{library} 

\end{document}